\documentclass[letterpaper, 10 pt, conference]{ieeeconf}

\IEEEoverridecommandlockouts
\title{\LARGE \bf
Trajectory-Consistent Flow Matching for Robust Visuomotor Policy Learning
}

\author{Riad Ahmed$^*$, Sujosh Nag, Moniruzzaman Akash, Mostafa Hussein, and Momotaz Begum%
\thanks{The authors are with the University of New Hampshire, Durham, NH, USA. Emails: \texttt{\{Riad.Ahmed, sujosh.nag, moniruzzaman.akash, Mostafa.Hussein, mbegum\}@unh.edu}.}%
\thanks{$^*$Corresponding author.}%
}

\usepackage{amsmath,amssymb}
\usepackage{booktabs}
\usepackage{graphicx}
\usepackage{xcolor}
\usepackage{cuted}
\usepackage{caption}
\usepackage{placeins}
\usepackage{balance}
\usepackage{microtype}

\AtBeginDocument{%
  \setlength{\abovedisplayskip}{4pt plus 2pt minus 2pt}%
  \setlength{\belowdisplayskip}{4pt plus 2pt minus 2pt}%
  \setlength{\abovedisplayshortskip}{1pt plus 1pt}%
  \setlength{\belowdisplayshortskip}{2pt plus 1pt minus 1pt}%
}

\let\oldthebibliography\thebibliography
\renewcommand{\thebibliography}[1]{%
  \oldthebibliography{#1}%
  \setlength{\itemsep}{0pt plus 0.5pt}%
  \setlength{\parskip}{0pt}%
}
\begin{document}
\maketitle
\thispagestyle{empty}
\pagestyle{empty}
\begin{strip}
  \centering
  \includegraphics[scale = 0.7]{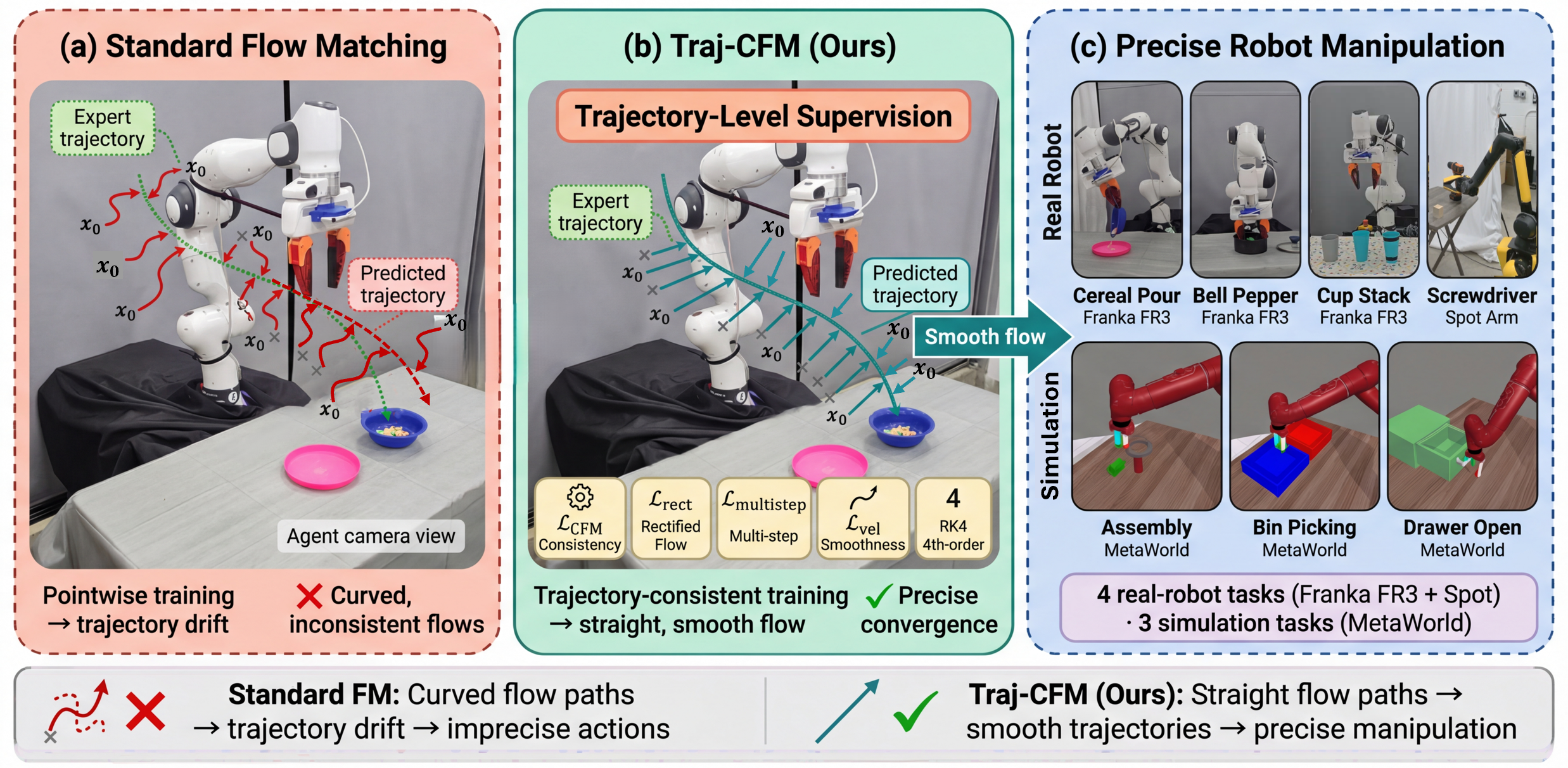}
  \captionof{figure}{Overview of Trajectory-Consistent Flow Matching (Traj-Consistent FM) for visuomotor
  policy learning. \textbf{(a)}~Standard flow matching trains pointwise velocity
  targets, producing curved flow paths from noise~$x_0$ to expert
  actions~$x_1$. The resulting ODE integration drifts from the expert trajectory,
  leading to imprecise robot actions. \textbf{(b)}~Our Traj-Consistent FM enforces
  trajectory-level supervision through four training losses
  ($\mathcal{L}_{\mathrm{rect}}$,
  $\mathcal{L}_{\mathrm{multistep}}$, $\mathcal{L}_{\mathrm{vel}}$, $\mathcal{L}_{\mathrm{action}}$)
  built atop the base $\mathcal{L}_{\mathrm{CFM}}$, combined with RK4
  inference, yielding straight, consistent flow paths that converge precisely to
  expert actions. \textbf{(c)}~Traj-Consistent FM enables precise manipulation across four
  real-robot tasks on Franka~FR3 and Boston Dynamics Spot, and three simulated
  MetaWorld tasks.}
  \label{fig:method}
\end{strip}
\thispagestyle{empty}
\pagestyle{empty}

%%%%%%%%%%%%%%%%%%%%%%%%%%%%%%%%%%%%%%%%%%%%%%%%%%%%%%%%%%%%%%%%%%%%%%%%%%%%%%%%

\begin{abstract}
Flow matching policies learn continuous velocity fields that transport noise to actions, enabling fast deterministic inference for robot manipulation. However, standard training optimizes a \emph{pointwise} velocity objective while inference requires numerical \emph{integration} of that field---a mismatch that causes compounding trajectory errors. We propose four complementary remedies: (1)~\emph{auxiliary rectified flow velocity regression} that provides uniform temporal supervision across the full time interval; (2)~\emph{multi-step trajectory consistency training} that supervises the integrated displacement of the velocity field over trajectory segments, directly closing the train--inference gap; (3)~\emph{velocity field regularization} that enforces temporal smoothness, preventing oscillations that destabilize integration; and (4)~\emph{fourth-order Runge-Kutta (RK4) inference} that reduces global discretization error by orders of magnitude over Euler methods. Critically, these components are not independently sufficient---RK4 without a smooth velocity field fails, and smoothness without trajectory-level supervision still drifts, as our ablation study confirms. We further pair these with a dual-view 3D point cloud encoder using two independent PointNet encoders for complementary spatial perception. On four real-robot tasks across a Franka arm and a Boston Dynamics Spot, our method achieves 70\% and 60\% overall success on two long-horizon multi-phase tasks where both baselines score 0\%, and reaches 100\% on precision tool placement. Three MetaWorld simulation tasks confirm consistent improvements, validating that trajectory-level supervision is essential for reliable policy execution.
\end{abstract}
%
%%%%%%%%%%%%%%%%%%%%%%%%%%%%%%%%%%%%%%%%%%%%%%%%%%%%%%%%%%%%%%%%%%%%%%%%%%%%%%%%
\section{INTRODUCTION}
Learning precision manipulation policies that preserve multimodality---different grasp orientations, approach trajectories, and timing variations---remains a central challenge in robot learning. Standard behavior cloning with regression losses averages over multiple valid modes, producing physically unrealizable actions \cite{pomerleau1989alvinn,florence2021ibc}. Diffusion policies \cite{chi2023diffusionpolicy,ze20243ddp} model the full action distribution via iterative denoising, but require 50--100 network evaluations per action query, limiting real-time deployment. Flow matching \cite{lipman2023flow,liu2023rectifiedflow,tong2024cfmot} provides a faster deterministic alternative: a learned velocity field transports noise to actions along a continuous ODE, and when the learned trajectories are straight, integration is both fast and accurate. Recent work pairing flow matching with 3D point cloud observations has further advanced manipulation performance \cite{zhang2025flowpolicy,noh20253d,chisari2024learning}.

Despite these advances, we identify a systematic \textbf{train--inference mismatch} that limits existing flow matching policies. During training, the velocity field is supervised at isolated, independently sampled time instances. At inference, however, the same field must be \emph{integrated sequentially}---each step's output becomes the next step's input. This gap manifests as three failure modes: (i)~\emph{collapsed paired-time supervision}, where the schedule-based pairing collapses most samples to a single fixed endpoint, leaving the velocity field without diverse inter-step consistency constraints across the time interval; (ii)~\emph{compounding integration drift}, where small per-step errors accumulate as the solver queries the network at states far from the training distribution; and (iii)~\emph{velocity oscillations} that satisfy pointwise objectives but destabilize numerical integration, causing jerky motions and large positional errors. These are especially damaging for precision manipulation.

Individually, partial remedies exist: rectified flow promotes straight trajectories, consistency training enforces local agreement, velocity regularization penalizes oscillations, and higher-order solvers reduce truncation error. However, no prior work addresses all three failure modes simultaneously in a unified framework. Critically, our ablation confirms the interaction is non-trivial: removing $\mathcal{L}_{\mathrm{vel}}$ while keeping RK4 drops overall success from 70\% to just 10\%, demonstrating that the benefit arises from the specific combination rather than any individual component.

We introduce \textbf{Trajectory-Consistent Flow Matching (Traj-Consistent FM)}, which combines four complementary components to close this gap: (1)~rectified flow velocity regression for uniform temporal coverage across $[0,1]$, (2)~multi-step trajectory consistency that directly supervises integrated displacement over trajectory segments, (3)~velocity smoothness regularization that prevents temporal oscillations, and (4)~RK4 inference that reduces global discretization error by orders of magnitude over Euler. We further design a dual-view 3D point cloud architecture with two independent PointNet encoders for complementary spatial perception.

\noindent Our contributions:
\begin{itemize}
  \item We identify a three-part train--inference gap in flow matching policies---collapsed paired-time supervision, integration drift, and velocity oscillation---and show all three must be addressed jointly.
  \item We introduce complementary training losses that produce a globally consistent, temporally smooth velocity field amenable to high-order integration.
  \item We deploy RK4 inference and a dual-view 3D point cloud architecture, ablating both against alternatives.
  \item On four real-robot tasks across Franka FR3 and Boston Dynamics Spot, we achieve 70\% and 60\% on long-horizon tasks where baselines score 0\%, and 100\% on precision tool placement.
\end{itemize}
%%%%%%%%%%%%%%%%%%%%%%%%%%%%%%%%%%%%%%%%%%%%%%%%%%%%%%%%%%%%%%%%%%%%%%%%%%%%%%%%
\section{RELATED WORK}
\subsection{Generative Policies for Robot Manipulation}

Imitation learning has evolved from regression-based cloning \cite{ross2011dagger,zhao2023act} and energy-based models \cite{florence2021ibc} to generative formulations. Diffusion Policy \cite{chi2023diffusionpolicy} models action distributions as denoising processes, with extensions to 3D point clouds \cite{ze20243ddp} and foundation-model scale \cite{black2024pi0}. Flow matching policies \cite{zhang2025flowpolicy,chisari2024learning,hu2024adaflow} replace stochastic diffusion with deterministic ODEs for faster inference, while ManiCM \cite{lu2024manicm} and Consistency Policy \cite{prasad2024consistencyfm} further compress function evaluations via consistency distillation. Both reduce inference cost but retain pointwise training and first-order solvers, addressing neither trajectory-level drift nor velocity field smoothness.

\subsection{Shortcomings of Flow Matching for Policy Learning}

Conditional Flow Matching \cite{lipman2023flow,albergo2023cfm} trains velocity networks at independent time points; the composed inference trajectory is never directly supervised. Rectified Flow \cite{liu2023rectifiedflow} and OT-CFM \cite{tong2024cfmot} promote straighter couplings but retain isolated-time supervision. Consistency Flow Matching \cite{yang2024consistencyflowmatching} enforces segment-boundary agreement without supervising integrated displacement. Multistep Consistency Models \cite{heek2024multistep} partition the ODE into segments yet still use first-order Euler.

TPC-FM \cite{li2024tpcfm} regularizes velocity oscillations through paired-time penalties, sharing motivation with our $\mathcal{L}_{\mathrm{vel}}$, but does not supervise integrated displacement or adopt higher-order solvers. CTM \cite{kim2024ctm} learns anytime-to-anytime jumps via teacher distillation from a pretrained score model, which does not accommodate the changing observation conditioning of closed-loop control. We supervise integrated displacements directly via $\mathcal{L}_{\mathrm{multistep}}$ without requiring a teacher. Flow Matching robotic policies \cite{zhang2025flowpolicy,hu2024adaflow,rouxel2024flowil} inherit all these gaps.

\subsection{Flow Matching Advances in Computer Vision}

InstaFlow \cite{liu2023instaflow} applies iterative reflow for one-step generation; Scaling Rectified Flow Transformers \cite{esser2024sd3} motivate higher-order integration. DPM-Solver \cite{lu2022dpmsolver} and DPM-Solver-v3 \cite{zheng2024dpmsolver3} replace Euler with high-order solvers, reducing truncation error by orders of magnitude. However, these solvers exploit noise-schedule structure (log-SNR reparameterization) specific to diffusion; adapting them to flow matching with observation-conditioned velocity fields would require non-trivial modifications. We use classical RK4, which assumes only smoothness of the velocity field. Our work is the first to bring trajectory-level supervision and higher-order Runge--Kutta integration together into a robot visuomotor policy.

%%%%%%%%%%%%%%%%%%%%%%%%%%%%%%%%%%%%%%%%%%%%%%%%%%%%%%%%%%%%%%%%%%%%%%%%%%%%%%%%
\section{PRELIMINARIES} \label{FMbasics}

Flow matching \cite{lipman2023flow,liu2023rectifiedflow} learns a velocity field transporting noise to data along a deterministic ODE. In visuomotor policy learning, the target is the conditional action distribution given observations and the source is standard Gaussian noise.

Given a noise sample $x_0 \sim \mathcal{N}(0, I)$ and an expert action $x_1$ from the demonstration dataset, flow matching constructs a linear interpolation path:
\begin{equation}
  x_t = (1 - t)\,x_0 + t\,x_1, \quad t \in [0, 1].
  \label{eq:interp}
\end{equation}
A neural network $v_\theta(x_t, t, c)$, conditioned on observation feature $c$, predicts the velocity along this path. For the linear path the ground-truth velocity is constant: $u^*(t) = x_1 - x_0$. At inference, actions are generated by drawing $x_0 \sim \mathcal{N}(0, I)$ and integrating the flow ODE:
\begin{equation}
  \frac{dx}{dt} = v_\theta(x, t, c), \quad x(0) = x_0, \quad \hat{x}_1 = x(1).
  \label{eq:ode}
\end{equation}

Consistency Flow Matching (Consistency-FM) \cite{yang2024consistencyflowmatching} enforces that predictions from nearby path points agree when projected to a shared segment boundary. The interval $[0,1]$ is partitioned into $K$ segments with boundaries $0 = s_0 < s_1 < \cdots < s_K = 1$; let $\tau(t) = s_k$ denote the endpoint of the segment containing~$t$. Training samples $t \sim \mathcal{U}[\varepsilon, 1]$ ($\varepsilon = 5 \times 10^{-3}$) and pairs each $t$ with $r = \min(t + \delta, 1)$ where $\delta$ is a schedule parameter. The loss requires Euler-forward steps from $x_t$ and $x_r$ to reach the same segment endpoint with matching velocities:
\begin{align}
  f_\theta(x,t,c) &= x + \bigl(\tau(t) - t\bigr)\,v_\theta(x,t,c),
  \label{eq:euler_step} \\
  \mathcal{L}_{\mathrm{CFM}} &=
    \mathbb{E}_{x_0,\,x_1,\,t}\!\Big[
      \bigl\| f_\theta(x_t,t,c) - f_\theta(x_r,r,c) \bigr\|^2 \nonumber\\
      &\quad+ \alpha\,\bigl\| v_\theta(x_t,t,c) - v_\theta(x_r,r,c) \bigr\|^2
    \Big],
  \label{eq:cfm}
\end{align}
where $\alpha = 0.8$. Deployment uses first-order Euler steps to integrate the learned velocity field. A gap remains between this pointwise training and the trajectory-level integration required at inference.
%
%%%%%%%%%%%%%%%%%%%%%%%%%%%%%%%%%%%%%%%%%%%%%%%%%%%%%%%%%%%%%%%%%%%%%%%%%%%%%%%%
\section{PROBLEM DEFINITION}
\label{sec:probdef}

We identify three gaps between the training objective of standard flow matching policies, as described in Section~\ref{FMbasics}, and the behavior required at inference time.

\paragraph{Collapsed paired-time supervision}
Consistency-FM pairs each $t$ with $r = \min(t + \delta, 1)$; with the default schedule parameter $\delta = 0.7$ \cite{yang2024consistencyflowmatching}, every $t > 0.3$ has $r$ clamped to~1. Consequently, no intermediate-to-intermediate pairs exist in the upper 70\% of $[0,1]$---the loss only constrains agreement between each such $t$ and the fixed endpoint $t{=}1$. Near $t = 1$ the pair separation $|r - t|$ vanishes, providing negligible gradient signal. The ODE integrator, however, must traverse the entire interval at inference, including regions where inter-step consistency is never directly enforced.

\paragraph{Trajectory integration drift}
The loss evaluates $v_\theta$ at points on the ground-truth path $x_t$, but at inference the solver chains predictions sequentially: after step~$n$ the state $x_{t_n}$ has drifted from the training path, and the network is queried at off-path states it was never trained on. This compounds across the trajectory and is most harmful in long-horizon tasks.

\paragraph{Integration instability}
No constraint prevents $v_\theta$ from oscillating rapidly in~$t$ while still satisfying the pairwise consistency condition. The first-order Euler integrator amplifies such oscillations, producing jerky motions and large positional errors.

All three gaps stem from the same root cause: training at isolated time points without feedback about the integrated trajectory. We address them jointly in Section~\ref{sec:method}.

%%%%%%%%%%%%%%%%%%%%%%%%%%%%%%%%%%%%%%%%%%%%%%%%%%%%%%%%%%%%%%%%%%%%%%%%%%%%%%%%
\section{METHOD}
\label{sec:method}
Section~\ref{FMbasics} established $\mathcal{L}_{\mathrm{CFM}}$~\eqref{eq:cfm} and the flow ODE~\eqref{eq:ode}; Section~\ref{sec:probdef} identified three gaps that arise when the pointwise-trained velocity field is integrated at inference. We present four training losses and one inference-time change that close these gaps.

% \vspace{0.5cm}
\begin{figure*}[t]
  \centering
  \vspace{0.5em}
  \includegraphics[width=0.7\textwidth]{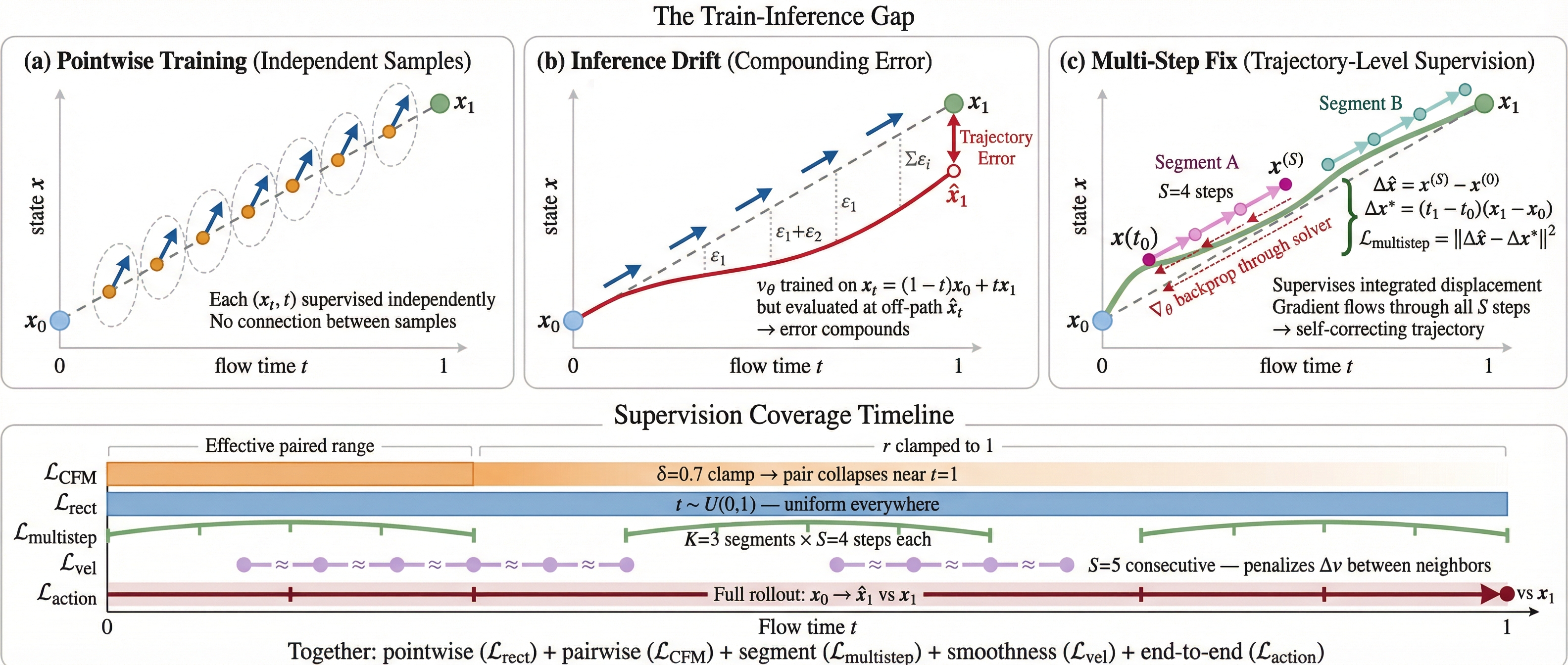}

\caption{\textbf{Train--inference gap in flow policies and our remedy.}
(a) Pointwise flow matching trains $v_\theta(x_t,t)$ at independently sampled times, without enforcing consistency along an integrated trajectory.
(b) During inference, the solver composes predictions sequentially; small per-step errors $\epsilon_i$ move the state off the training path and compound, yielding $\hat{x}_1 \neq x_1$.
(c) Our trajectory-consistency objectives supervise integrated displacements over short multi-step segments and through rollouts, encouraging a self-correcting velocity field.
\textbf{Bottom:} supervision coverage over $t\in[0,1]$ for each loss component.}

  \label{fig:gap}
\end{figure*}

\subsection{Auxiliary Rectified Flow Velocity Regression}
\label{sec:rectflow}

To fill the coverage gaps left by the schedule-based Consistency-FM loss, we add a regression loss that draws $t$ uniformly from $[0, 1]$ and supervises the velocity field against its analytical target \cite{liu2023rectifiedflow}. For the linear path, \eqref{eq:interp}, the target is constant:
\begin{equation}
  u^*(t) = x_1 - x_0, \quad \forall\, t \in [0, 1].
  \label{eq:rect_vel}
\end{equation}
Any temporal variation in $v_\theta$ introduces curvature that increases the number of function evaluations (NFEs) required for accurate integration. We minimize:
\begin{equation}
  \mathcal{L}_{\mathrm{rect}} =
    \mathbb{E}_{x_0,\,x_1,\,t \sim \mathcal{U}(0,1)}\!\left[
      \bigl\| v_\theta(x_t, t, c) - (x_1 - x_0) \bigr\|^2
    \right].
  \label{eq:rect_loss}
\end{equation}
This provides dense supervision across $[0,1]$, penalizes $t$-dependence, and straightens learned flows. We set $\lambda_{\mathrm{rect}} = 1.0$.

\subsection{Multi-Step Trajectory Consistency}
\label{sec:multistep}

Even a pointwise-accurate velocity field can drift when integrated, because errors shift the state off the training distribution. We supervise the \emph{outcome of integration} over trajectory segments. For a randomly sampled pair $t_0 < t_1$ drawn from $[0, 1]$, the ground-truth
displacement along path \eqref{eq:interp} from $t_0$ to $t_1$ is:
\begin{equation}
  \Delta x^* = (t_1 - t_0)(x_1 - x_0).
  \label{eq:true_delta}
\end{equation}
Starting from the on-path point $x^{(0)} = (1 - t_0)x_0 + t_0 x_1$, we roll out
the learned velocity field using $S = 4$ Euler steps with step size $h = (t_1 - t_0)/S$:
\begin{equation}
  x^{(s+1)} = x^{(s)} + h\,v_\theta\!\left(x^{(s)},\;
    t_0 + s\,h,\; c\right),
  \label{eq:seg_rollout}
\end{equation}
and require the predicted displacement $\hat{\Delta x} = x^{(S)} - x^{(0)}$ to
match the ground truth:
\begin{equation}
  \mathcal{L}_{\mathrm{multistep}} =
    \mathbb{E}_{x_0,\,x_1,\,t_0,\,t_1}\!\left[
      \bigl\| \hat{\Delta x} - \Delta x^* \bigr\|^2
    \right].
  \label{eq:cons_loss}
\end{equation}
We sample $K = 3$ independent segments per training step and average their losses. The gradient backpropagates through all $S$ steps, training $v_\theta$ to produce mutually consistent sequences---unlike $\mathcal{L}_{\mathrm{CFM}}$ and $\mathcal{L}_{\mathrm{rect}}$, which see each time step in isolation. The multi-step rollout~\eqref{eq:seg_rollout} is a composition $x^{(S)} = g_S \circ \cdots \circ g_1(x^{(0)})$, where $g_s(x) \triangleq x + v_\theta(x,\,t_0 + \tfrac{s}{S}(t_1-t_0),\,c)\,\tfrac{t_1-t_0}{S}$ is a single Euler step and $c$ is produced by the observation encoder. By the chain rule, $\partial \mathcal{L}_{\mathrm{multistep}}/\partial c$ accumulates gradient contributions from all $S$ steps. Concretely, if the encoder confuses two visually similar objects (e.g., two cups differing only by a thin stripe), the conditioning vector $c$ shifts by $\delta c$; each step then produces a velocity error $\delta v \propto \partial v_\theta/\partial c \cdot \delta c$, and after $S$ steps the displacement error grows as $\|\hat{\Delta x} - \Delta x^*\| \propto S \cdot \|\delta v\| \cdot \Delta t$. By contrast, $\mathcal{L}_{\mathrm{CFM}}$ and $\mathcal{L}_{\mathrm{rect}}$ evaluate a single network call per sample, so the encoder gradient reflects only a one-step velocity deviation---too weak to penalize subtle encoding errors. The $S$-step amplification in $\mathcal{L}_{\mathrm{multistep}}$ thus produces a gradient signal proportional to the \emph{integrated} cost of misidentification, making the encoder sensitive to fine-grained visual distinctions that single-step losses ignore.

\subsection{Velocity Smoothness Regularization}
\label{sec:velreg}

The Consistency-FM loss places no constraint on how $v_\theta$ varies between paired times, allowing temporally irregular fields that degrade RK4's four intermediate evaluations. We penalize consecutive velocity differences along $S = 5$ uniformly spaced times on a random path:
\begin{equation}
  \mathcal{L}_{\mathrm{vel}} =
    \mathbb{E}\!\left[\,\frac{1}{S-1}
      \sum_{i=1}^{S-1}
        \bigl\| v_\theta(x_{t_{i+1}}, t_{i+1}, c)
               - v_\theta(x_{t_i}, t_i, c) \bigr\|^2
    \right].
  \label{eq:smooth}
\end{equation}
We omit a magnitude penalty $\|v_\theta\|^2$ since the target $u^* = x_1 - x_0$ has non-trivial magnitude. $\mathcal{L}_{\mathrm{rect}}$ anchors the field to the correct constant target; $\mathcal{L}_{\mathrm{vel}}$ penalizes variation around it.

\subsection{RK4 Integration at Inference}
\label{sec:rk4}

At inference we replace the standard Euler method with fourth-order Runge-Kutta
(RK4) \cite{butcher2016runge} as the numerical ODE solver for integrating the
flow ODE \eqref{eq:ode}. RK4 queries the conditional vector field at four
evaluation points within the current time step $[t_n, t_n + \Delta t]$:
\begin{align}
  k_1 &= v_\theta(x_n,\; t_n,\; c)\,\Delta t, \label{eq:k1} \\
  k_2 &= v_\theta\!\left(x_n + \tfrac{k_1}{2},\; t_n + \tfrac{\Delta t}{2},\; c\right)\Delta t, \label{eq:k2} \\
  k_3 &= v_\theta\!\left(x_n + \tfrac{k_2}{2},\; t_n + \tfrac{\Delta t}{2},\; c\right)\Delta t, \label{eq:k3} \\
  k_4 &= v_\theta(x_n + k_3,\; t_n + \Delta t,\; c)\,\Delta t, \label{eq:k4} \\
  x_{n+1} &= x_n + \tfrac{1}{6}(k_1 + 2k_2 + 2k_3 + k_4). \label{eq:rk4_update}
\end{align}
RK4 achieves $\mathcal{O}(\Delta t^4)$ global truncation error versus $\mathcal{O}(\Delta t)$ for Euler, where each RK4 step requires 4~NFEs (network forward passes). We deploy $N{=}30$ RK4 steps ($120$~NFEs); Fig.~\ref{fig:gap} visualizes the improvement.

At a fixed NFE budget $M$, Euler takes $M$ steps at $\Delta t_E = 1/M$ while RK4 takes $M/4$ steps at $\Delta t_R = 4/M$, both using exactly $M\tau$ wall-clock time ($\tau$ = one forward pass). Global truncation error scales as $\varepsilon_{\mathrm{Euler}} = C_1 \Delta t_E$ and $\varepsilon_{\mathrm{RK4}} = C_4 \Delta t_R^{\,4}$, where $C_1, C_4$ depend on the regularity of the velocity field. Their ratio is $\varepsilon_{\mathrm{Euler}}/\varepsilon_{\mathrm{RK4}} = C_1\,\Delta t_E \,/\, (C_4\,\Delta t_R^{\,4})$. With $\Delta t_R = 4\,\Delta t_E$ this becomes $(C_1/C_4)\cdot\Delta t_E^{-3}/256$, i.e.\ the advantage grows as $\mathcal{O}(\Delta t_E^{-3})$ with the budget. Assuming comparable constants ($C_1 \approx C_4$), our budget $M{=}120$ gives $\varepsilon_{\mathrm{Euler}}/\varepsilon_{\mathrm{RK4}} \approx 6{,}750$; Euler would need ${\sim}810{,}000$ steps to match---far exceeding any real-time budget at 20\,Hz.
Critically, RK4's advantage requires a smooth velocity field: the four evaluations $k_1$--$k_4$ per step must return consistent estimates, which oscillatory fields violate. Dedicated diffusion solvers such as DPM-Solver \cite{lu2022dpmsolver} exploit log-SNR reparameterization specific to score-based models and do not directly apply to observation-conditioned flow matching ODEs. 
\begin{figure}[t]
  \centering
  \includegraphics[width=\linewidth]{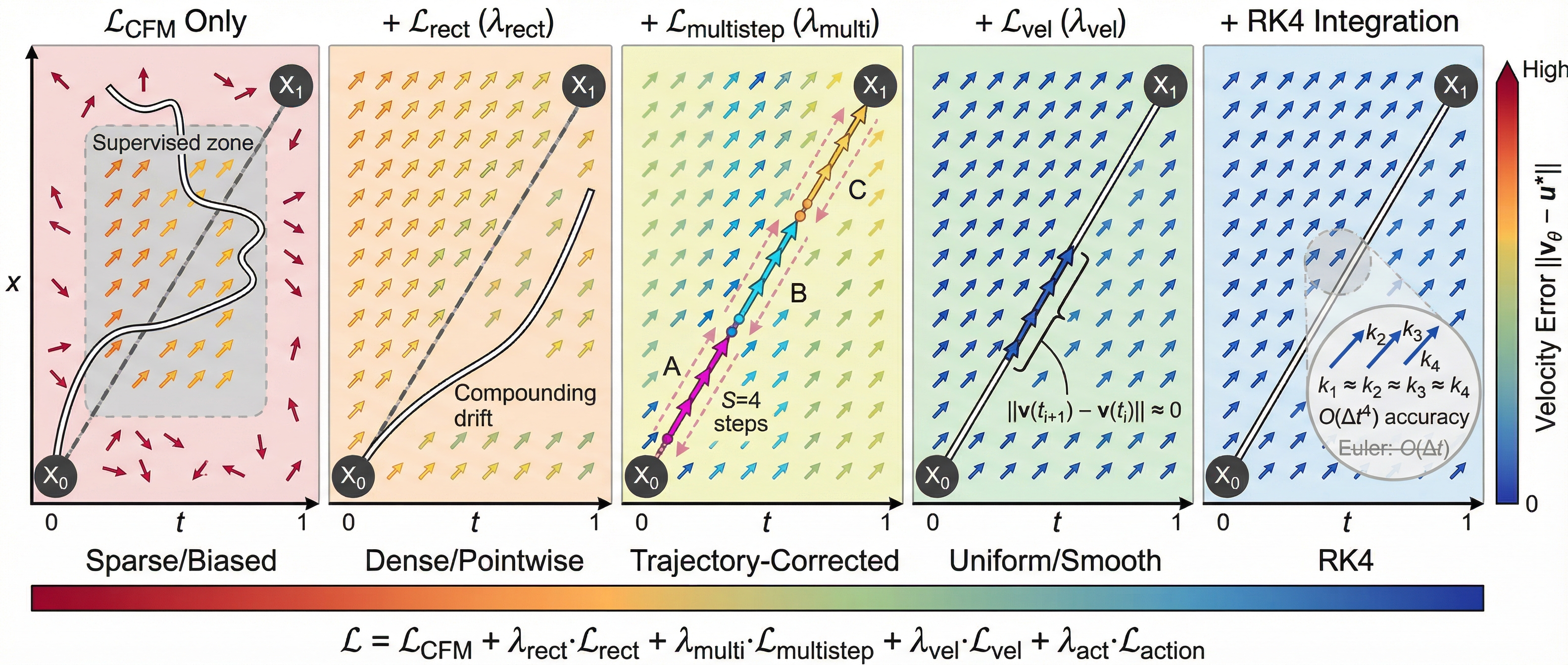}
%   \caption{Progressive effect of each loss component on the learned velocity field, visualized as arrow plots in the $(t, x)$ plane with color indicating velocity error $\|v_\theta - u^*\|$.
%   \textbf{(1)}~$\mathcal{L}_{\mathrm{CFM}}$ alone produces a sparse, biased supervised zone with large unsupervised regions.
%   \textbf{(2)}~Adding $\mathcal{L}_{\mathrm{rect}}$ provides dense, pointwise supervision across the full $[0,1]$ interval, but compounding drift remains.
%   \textbf{(3)}~$\mathcal{L}_{\mathrm{multistep}}$ corrects trajectory-level drift by supervising $S{=}4$-step integrated segments (A$\to$B$\to$C), aligning predicted displacements with ground truth.
%   \textbf{(4)}~$\mathcal{L}_{\mathrm{vel}}$ enforces temporal smoothness, making consecutive velocities nearly identical ($\|v(t_{i+1}) - v(t_i)\| \approx 0$) and yielding a uniform, low-error field.
%   \textbf{(5)}~At inference, RK4 exploits the smooth field: its four evaluations $k_1 \approx k_2 \approx k_3 \approx k_4$ achieve $\mathcal{O}(\Delta t^4)$ accuracy, far surpassing Euler's $\mathcal{O}(\Delta t)$.
% }
\caption{Vector fields visualize the learned velocity $v_\theta(t,x)$ in a 1D toy $(t,x)$ setting; color encodes $\lVert v_\theta-u^*\rVert$.  (a) $\mathcal{L}_{CFM}$ alone provides sparse supervision and induces bias. (b) $\mathcal{L}_{rect}$ adds dense pointwise constraints but does not eliminate compounding drift. (c) $\mathcal{L}_{multistep}$ constrains integrated $S{=}4$ step displacements, reducing drift. (d) $\mathcal{L}_{vel}$ penalizes temporal variation in $v_\theta$, yielding smoother, more uniform fields. (e) At inference, RK4 benefits from the smoother field and exhibits lower discretization error than Euler.}
  \label{fig:losses}
\end{figure}
\subsection{Full Training Objective}
\label{sec:objective}

The complete training objective is:
\begin{equation}
  \mathcal{L} =
    \mathcal{L}_{\mathrm{CFM}}
    + \lambda_{\mathrm{r}}\,\mathcal{L}_{\mathrm{rect}}
    + \lambda_{\mathrm{c}}\,\mathcal{L}_{\mathrm{multistep}}
    + \lambda_{\mathrm{v}}\,\mathcal{L}_{\mathrm{vel}}
    + \lambda_{\mathrm{a}}\,\mathcal{L}_{\mathrm{action}},
  \label{eq:total_loss}
\end{equation}
where $\mathcal{L}_{\mathrm{action}} = \mathbb{E}[\|\hat{x}_1 - x_1\|^2]$ runs a $S_{\mathrm{act}}{=}5$-step Euler rollout from $x_0 \sim \mathcal{N}(0,I)$ to $t{=}1$ and penalizes endpoint error. This uses fewer steps and a simpler solver than the 30-step RK4 at deployment; back-propagating through 120~NFEs would be memory-prohibitive. The mismatch is benign: $\mathcal{L}_{\mathrm{action}}$ acts as a coarse end-to-end sanity check ($\lambda_{\mathrm{a}}{=}0.1$), while the gap-closing losses $\mathcal{L}_{\mathrm{multistep}}$ and $\mathcal{L}_{\mathrm{vel}}$ supervise the velocity field directly and are solver-agnostic.

The five losses provide complementary supervision at different levels: $\mathcal{L}_{\mathrm{CFM}}$ enforces local pairwise consistency, $\mathcal{L}_{\mathrm{rect}}$ anchors each prediction to the ground-truth velocity, $\mathcal{L}_{\mathrm{multistep}}$ supervises the integrated trajectory, $\mathcal{L}_{\mathrm{vel}}$ regularizes temporal smoothness, and $\mathcal{L}_{\mathrm{action}}$ checks the final output. We set $\lambda_{\mathrm{r}} = 1.0$, $\lambda_{\mathrm{c}} = 0.5$, $\lambda_{\mathrm{v}} = 0.1$, $\lambda_{\mathrm{a}} = 0.1$. The higher weights on $\mathcal{L}_{\mathrm{CFM}}$ and $\mathcal{L}_{\mathrm{rect}}$ reflect their role as primary accuracy-driving losses, while the lower weights keep the regularizers subordinate. Figure~\ref{fig:losses} shows the convergence dynamics of each component.

\subsection{Observation Encoding}
\label{sec:obsenc}

Each observation consists of an agent-view point cloud $\mathcal{P}_a \in \mathbb{R}^{N \times 6}$ (XYZ+RGB) (where $N$ is the pointcloud number) an eye-in-hand point cloud $\mathcal{P}_e \in \mathbb{R}^{N \times 6}$, and proprioception $s \in \mathbb{R}^{21}$. Two independent PointNet encoders \cite{qi2017pointnet} with identical architecture but separate weights produce:
\begin{align}
  f_a &= \mathrm{PN}_{\theta_a}(\mathcal{P}_a) \in \mathbb{R}^{64}, \\
  f_e &= \mathrm{PN}_{\theta_e}(\mathcal{P}_e) \in \mathbb{R}^{64}.
\end{align}
The proprioceptive state gives $f_s = \mathrm{MLP}(s) \in \mathbb{R}^{64}$; the full observation vector is $c = [f_a;\; f_e;\; f_s] \in \mathbb{R}^{192}$.
Separate encoders let each view specialize: the agent-view encoder for object identity and global position, the eye-in-hand encoder for contact geometry and local detail.
The policy receives $n_{\mathrm{obs}} = 2$ consecutive steps; the conditioning signal is $C =[c_1;\; c_2] \in \mathbb{R}^{384}$.

\begin{figure*}[t]
  \centering
  \vspace{0.5em}
  \includegraphics[width=0.75\textwidth]{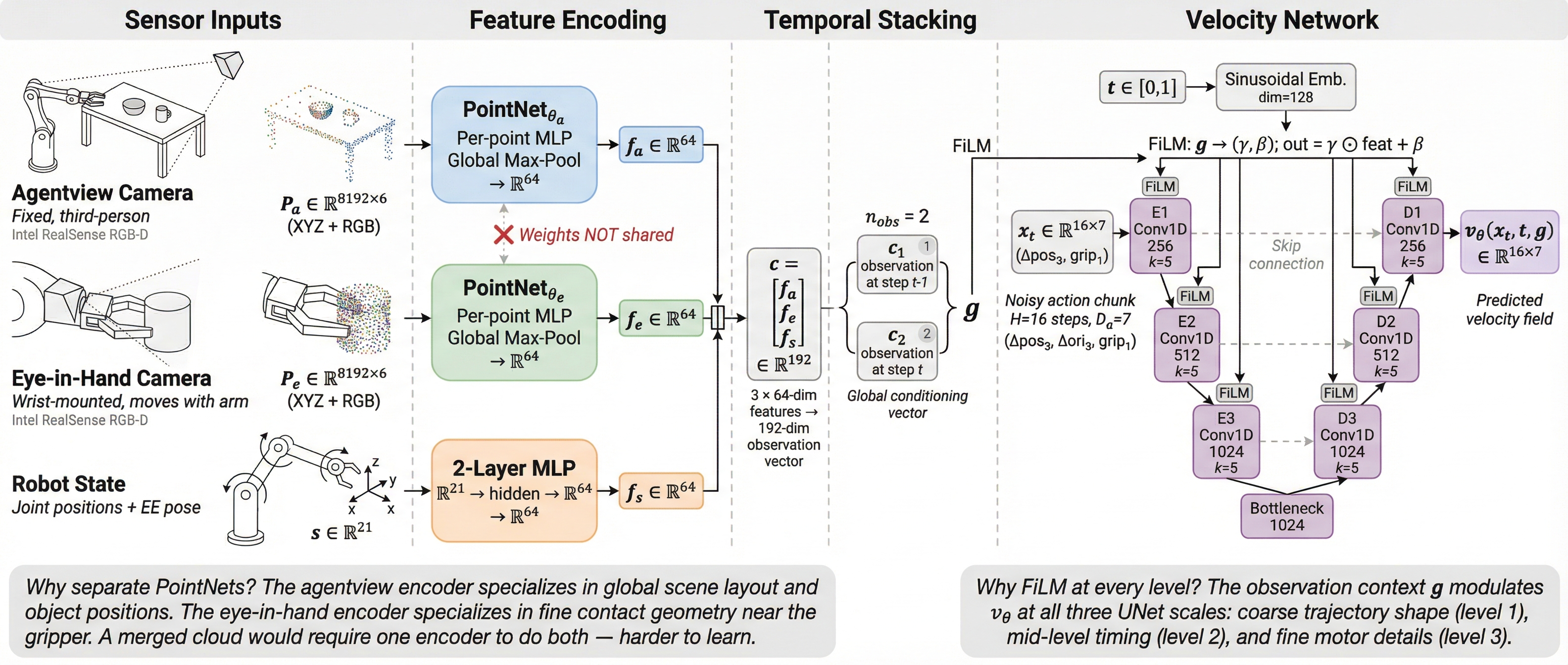}
  \caption{Dual-view 3D point cloud observation encoder.
  A fixed \textbf{agent-view} camera captures global object positions and
  workspace layout; a wrist-mounted \textbf{eye-in-hand} camera captures
  contact geometry around the gripper.
  Independent PointNet encoders $\mathrm{PN}_{\theta_a}$,
  $\mathrm{PN}_{\theta_e}$ extract 64-d features from each cloud.
  Features are concatenated with a 64-d proprioceptive embedding
  to form the 192-d conditioning vector $c$, stacked across two
  timesteps into the 384-d input $g$ for the velocity network.}
  \label{fig:dualview}
\end{figure*}

\subsection{Velocity Network Architecture}
\label{sec:arch}

The velocity network $v_\theta$ is a 1D conditional UNet \cite{chi2023diffusionpolicy} that operates along the temporal dimension of the action sequence $x_t \in \mathbb{R}^{H \times D_a}$, where $H=16$ is the prediction horizon (number of future action steps) and $D_a=7$ is the action dimension (3D position delta, 3D orientation delta, gripper command). The network receives three inputs: the noisy action sequence $x_t$, a sinusoidal time embedding (dim~128) encoding the flow time $t$, and the observation conditioning vector $C$, and outputs a velocity field of the same shape $\mathbb{R}^{H \times D_a}$. The UNet has three resolution levels with channel dimensions $(256, 512, 1024)$, kernel size~5, and FiLM conditioning \cite{perez2018film} that modulates intermediate features based on the time embedding and observation at each level. All parameters---both PointNet encoders and the UNet---are trained jointly end-to-end. We maintain an exponential moving average (EMA) of network weights with decay $0.9999$ and use the EMA model at deployment. Figure~\ref{fig:dualview} illustrates the full pipeline from point cloud observations to velocity field prediction.

\subsection{Training Configuration}
\label{sec:training}

All four auxiliary losses operate exclusively during training; the deployed inference pipeline differs from the baseline only by replacing Euler with RK4, adding zero architectural overhead (Section~\ref{sec:rk4}). The main training cost is $\mathcal{L}_{\mathrm{multistep}}$, requiring $S{=}4$ extra forward passes per segment with backpropagation through the solver, increasing per-epoch time by ${\sim}35\%$ and GPU memory by ${\sim}20\%$ (RTX 4090, batch~64; total: ${\sim}5.5$h vs.\ ${\sim}4$h for 400 epochs on Bell Pepper).

%%%%%%%%%%%%%%%%%%%%%%%%%%%%%%%%%%%%%%%%%%%%%%%%%%%%%%%%%%%%%%%%%%%%%%%%%%%%%%%%
\section{EXPERIMENTS}
\label{sec:experiments}

\subsection{Robot Platforms and Setup}
We evaluate on a \textbf{Franka Emika FR3} 7-DOF arm with parallel gripper and a \textbf{Boston Dynamics Spot} mobile manipulator (6-DOF arm). Both use calibrated RGB-D sensing from agent-view and wrist-view perspectives (Intel RealSense cameras on Franka; RealSense + onboard hand camera on Spot). Observations consist of two 8{,}192-point clouds (XYZ+RGB) and 21-d proprioception; actions are 7-d end-effector delta commands. Demonstrations are collected at 20\,Hz via SpaceMouse (Franka) and Meta Quest~3 teleoperation (Spot).

\subsection{Tasks}

We evaluate on four real-robot manipulation tasks across two platforms (Fig.~\ref{fig:real_tasks}). We group tasks into \emph{short-horizon} settings with a single continuous manipulation phase and \emph{long-horizon} settings that require executing multiple sequential subtasks with distinct phases.

\subsubsection{Short-Horizon Tasks}
\textbf{Cereal Pouring (Franka, 30 demos).} The robot grasps a bowl containing cereal, moves above a target plate, and pours. Due to high friction and granular jamming, success often requires a smooth tilting motion combined with small shaking perturbations to reliably empty the bowl.

\textbf{Screwdriver Placement (Spot, 32 demos).} Spot grasps a screwdriver and brings its tip into contact with a small target cube. Poses are randomized, emphasizing precise tool-tip alignment on a mobile manipulator.

\subsubsection{Long-Horizon Tasks}
\textbf{Bell Pepper Placing (Franka, 70 demos).} Two subtasks: (1) grasp a bell pepper and place it into a pot, then (2) grasp the lid and place it onto the pot with rim alignment. Demonstrations are collected across widely varying initial configurations---pepper, pot, and lid poses are randomized each episode---producing a broad action distribution that the policy must generalize over.
\textbf{Cup Stacking (Franka, 101 demos).} Two subtasks with three cups at randomized poses: (1) insert the plain blue cup into the gray cup, then (2) insert the black-striped blue cup into the assembled stack. The 101 demonstrations cover diverse spatial arrangements. This task couples precise sequential insertions with fine-grained visual discrimination between two nearly identical cups differing only by a thin black stripe.

\subsubsection{Simulation Tasks}
We additionally evaluate on three MetaWorld \cite{yu2020metaworld} tasks---\textbf{Drawer Open}, \textbf{Assembly} (peg insertion), and \textbf{Bin Picking}---which provide deterministic dynamics and noise-free observations. Since all methods reach near-perfect success, we use these tasks primarily to compare trajectory quality via the MetaWorld reward signal, which penalizes inefficient or jerky motions.

\begin{figure}[t]
  \centering
  \includegraphics[width=\linewidth]{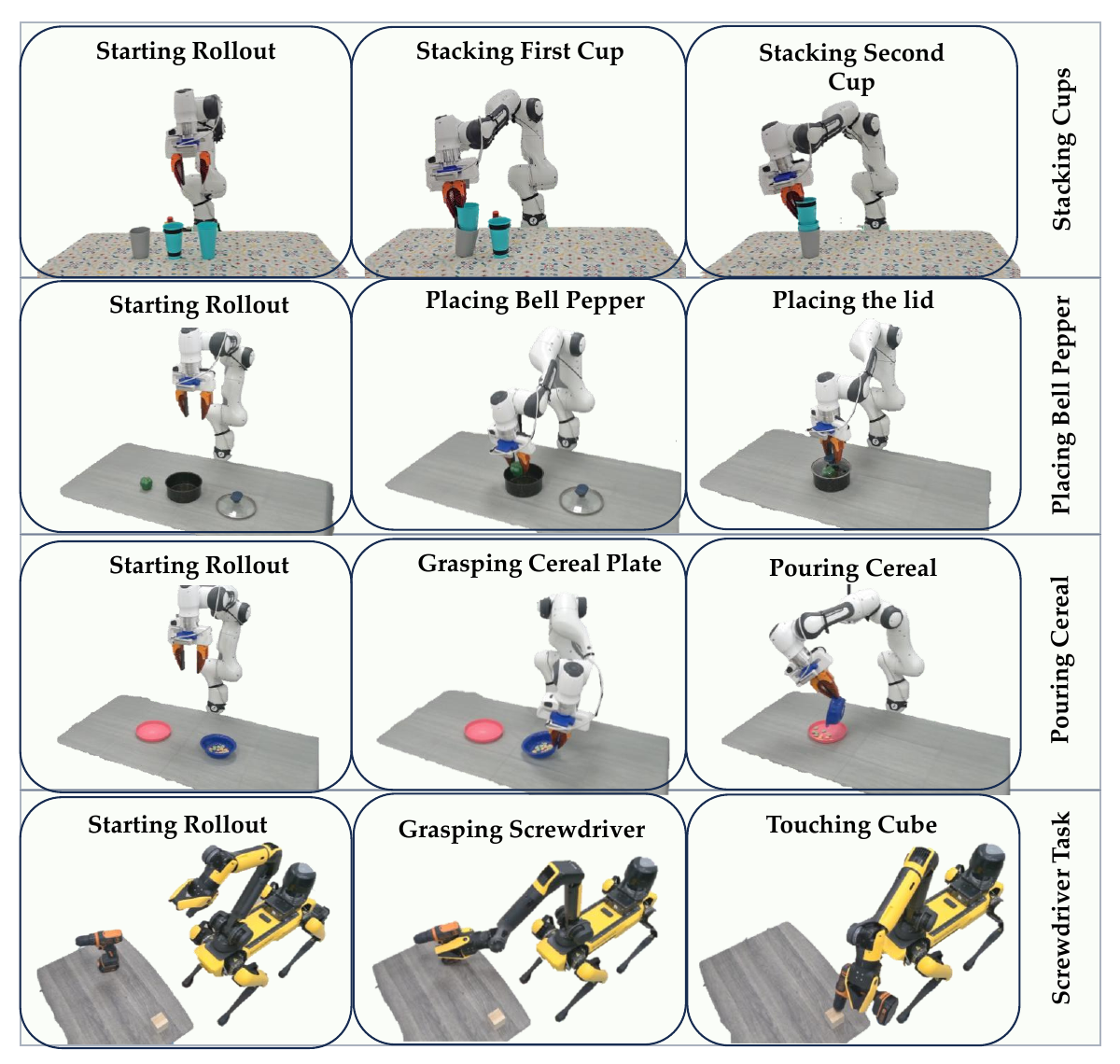}
  \caption{Real-robot task setups and execution sequences. \textbf{Top two rows:} Cup Stacking and Bell Pepper Placing on the Franka~FR3 arm.
  \textbf{Bottom two rows:} Cereal Pouring on the Franka~FR3 and Screwdriver
  Placement on the Boston Dynamics Spot. Each sequence shows key phases
  of the manipulation trajectory from approach to task completion.}
  \label{fig:real_tasks}
\end{figure}

\subsection{Baselines}

To ensure a fair comparison, all three methods share the same dual-view PointNet observation encoder (Section~\ref{sec:obsenc}) and the same 1D conditional UNet architecture (Section~\ref{sec:arch}); only the training objective and inference solver differ. This isolates the effect of our proposed losses and integration method from architectural choices.

\textbf{DP3} \cite{ze20243ddp}: A DDPM-based diffusion policy using 20-step DDIM denoising, representing the state-of-the-art diffusion paradigm for 3D point cloud policies.

\textbf{Consistency FM} \cite{zhang2025flowpolicy}: The FlowPolicy baseline with $\mathcal{L}_{\mathrm{CFM}}$ and first-order Euler integration---the base upon which Traj-Consistent FM is built, without our additional losses or RK4. This directly measures the contribution of our improvements.

\subsection{Simulation Results}

Table~\ref{tab:sim_results} reports success rate and mean trajectory reward
across three MetaWorld simulation tasks \cite{yu2020metaworld}, evaluated over
50 episodes per method.

\begin{table}[t]
\centering
\small
\setlength{\tabcolsep}{3pt}
\caption{Simulation results on three MetaWorld tasks (50 episodes each).
Columns per task: $S$ (\%) = Success Rate and $R$ = Reward.
Best per task in \textbf{bold}.}
\label{tab:sim_results}
\resizebox{\columnwidth}{!}{%
\begin{tabular}{l cc cc cc}
\toprule
& \multicolumn{2}{c}{Drawer Open}
& \multicolumn{2}{c}{Assembly}
& \multicolumn{2}{c}{Bin Picking} \\
\cmidrule(lr){2-3} \cmidrule(lr){4-5} \cmidrule(lr){6-7}
Method
  & $S$ & $R$
  & $S$ & $R$
  & $S$ & $R$ \\
\midrule
DP3 \cite{ze20243ddp}
  & 98.0 & 376.3
  & 88.0 & 241.6
  & 96.0 & 88.3 \\
Consistency FM \cite{zhang2025flowpolicy}
  & 100.0 & 386.6
  & 98.0 & 259.3
  & 96.0 & 88.5 \\
\textbf{Traj-Consistent FM (Ours)}
  & \textbf{100.0} & \textbf{446.3}
  & \textbf{98.0} & \textbf{281.2}
  & \textbf{96.0} & \textbf{89.5} \\
\bottomrule
\end{tabular}}
\end{table}

All three methods achieve near-perfect success rates, as expected with deterministic dynamics and noise-free observations. However, trajectory reward---penalizing inefficient or jerky motions---reveals meaningful quality differences. Traj-Consistent FM achieves the highest rewards across all tasks (e.g., 446.3 vs.\ 386.6 on Drawer Open, 15\% improvement), indicating smoother, more direct trajectories. This quality gap becomes performance-critical on physical robots, where noisy observations amplify integration errors.

\subsection{Short-Horizon Real-Robot Results}

Table~\ref{tab:real_short} reports success rates on the two
short-horizon real-robot tasks, evaluated over 20 trials per method.

\begin{table}[t]
\centering
\caption{Short-horizon real-robot results (success rate \%).
Best per task in \textbf{bold}. Evaluated over 20 trials.}
\label{tab:real_short}
\begin{tabular}{l cc}
\toprule
Method & Cereal Pouring & Screwdriver \\
\midrule
DP3 \cite{ze20243ddp}           & 55  & 70 \\
Consistency FM \cite{zhang2025flowpolicy} & 65  & 60 \\
\textbf{Traj-Consistent FM (Ours)}       & \textbf{80}  & \textbf{100} \\
\bottomrule
\end{tabular}
\end{table}

Traj-Consistent FM outperforms both baselines on both tasks. On Cereal Pouring (80\% vs.\ 65\%/55\%), $\mathcal{L}_{\mathrm{vel}}$ is the key differentiator: pouring requires a smooth tilting arc, and baseline failures involve abrupt wrist rotations that scatter cereal or fail to initiate pouring. On Screwdriver Placement (100\% vs.\ 60\%/70\%), $\mathcal{L}_{\mathrm{rect}}$ + RK4 delivers the endpoint precision needed for tool-tip contact with the small target cube; baseline failures stem from accumulated integration drift causing the tip to miss by several millimeters.

\subsection{Long-Horizon Real-Robot Results}

Table~\ref{tab:real_long} reports subtask-level and overall success
rates on the two long-horizon real-robot tasks. Each task consists of
two sequential subtasks; the trial proceeds to the second subtask only
if the first succeeds. Evaluation uses 20 trials per method.

\vspace{0.5cm}
\begin{table}[t]
\centering
\small
\setlength{\tabcolsep}{3pt}
\caption{Long-horizon real-robot results (success rate \%). S1 and S2
denote subtask~1 and subtask~2 success. Overall requires both
subtasks to succeed. Best per column in \textbf{bold}. Evaluated over
20 trials.}
\label{tab:real_long}
\resizebox{\columnwidth}{!}{%
\begin{tabular}{l ccc ccc}
\toprule
& \multicolumn{3}{c}{Bell Pepper Placing}
& \multicolumn{3}{c}{Cup Stacking} \\
\cmidrule(lr){2-4} \cmidrule(lr){5-7}
Method & S1 & S2 & Overall & S1 & S2 & Overall \\
\midrule
DP3 \cite{ze20243ddp}
  & 60 & 0 & 0
  & 55 & 0 & 0 \\
Consistency FM \cite{zhang2025flowpolicy}
  & 50 & 0 & 0
  & 40 & 0 & 0 \\
\textbf{Traj-Consistent FM (Ours)}
  & \textbf{95} & \textbf{70} & \textbf{70}
  & \textbf{90} & \textbf{60} & \textbf{60} \\
\bottomrule
\end{tabular}}
\end{table}

Neither baseline completed a single full trial (0\% overall) on either task, despite achieving 40--60\% S1 success---demonstrating that the baselines learn competent single-phase policies but cannot sustain accuracy through multi-phase rollouts. The broad configuration diversity in demonstrations amplifies this effect: the velocity field must remain accurate across a wide observation distribution, and pointwise-trained fields inevitably drift when queried at states accumulated over extended integration. Traj-Consistent FM achieves 95\%/70\% (S1/Overall) on Bell Pepper and 90\%/60\% on Cup Stacking because $\mathcal{L}_{\mathrm{multistep}}$ directly supervises integrated displacements across this configuration space, while $\mathcal{L}_{\mathrm{vel}}$ ensures the smooth field that RK4 requires for its theoretical accuracy advantage.

\paragraph{Bell Pepper Placing} Baselines fail through three recurring modes (Fig.~\ref{fig:real_tasks}): (a)~the policy confuses the pepper with the lid (object discrimination error), (b)~the arm drifts toward the wrong receptacle after grasping (placement target confusion), and (c)~the policy stalls after subtask~1 and hovers instead of transitioning to lid grasping. These failures are characteristic of accumulated integration drift: the velocity field produces reasonable per-step actions, but compounded errors shift the state to regions where the observation conditioning no longer specifies the correct subtask. Traj-Consistent FM's trajectory-level supervision prevents this by training the velocity network to maintain accurate displacements over extended horizons, keeping the integrated state within the learned distribution.

\paragraph{Cup Stacking} The two blue cups differ only by a thin black strip, making this task a direct test of the visual encoding mechanism described in Section~\ref{sec:multistep}: a single-step loss penalizes only the instantaneous velocity deviation from confusing the cups, whereas $\mathcal{L}_{\mathrm{multistep}}$'s $S$-step rollout amplifies that encoding error into a displacement loss proportional to the segment length. Baselines, trained with only pointwise losses, frequently pick the wrong cup first or produce jerky insertions that knock over the stack. Traj-Consistent FM makes zero cup-discrimination errors across all 20 evaluation trials, consistent with the $S{\times}$ gradient amplification driving the encoder to resolve fine-grained visual differences. Remaining failures occur only when cups are placed very close together, causing point cloud overlap. The 30-point S1-to-Overall drop reflects the compounding difficulty of the second insertion, where the target receptacle (the gray cup with first blue cup inside) presents a narrower opening.

\subsection{Ablation Study}

We ablate on Bell Pepper Placing---the longest-horizon task, where compounding errors are most visible.
Table~\ref{tab:ablation_pepper} reports success rates when individual components are removed.

\begin{table}[t]
\centering
\small
\setlength{\tabcolsep}{4pt}
\caption{Ablation study on Bell Pepper Placing (success rate \%).
Each row removes one component from the full Traj-Consistent FM model. Best per column in \textbf{bold}. Evaluated over
20 trials.}
\label{tab:ablation_pepper}
\begin{tabular}{l ccc}
\toprule
Configuration & S1 & S2 & Overall \\
\midrule
Full model (Traj-Consistent FM)            & \textbf{95} & \textbf{70} & \textbf{70} \\
w/o multi-step consistency loss  & 60  & 20  & 20  \\
w/o velocity regularization      & 50  & 10  & 10  \\
w/o RK4 (Euler inference)        & 80  & 40  & 40  \\
w/o dual-view (shared PointNet)  & 85  & 45  & 45  \\
\bottomrule
\end{tabular}
\end{table}

Every component contributes independently, and the magnitudes of the drops are informative.
Removing $\mathcal{L}_{\mathrm{vel}}$ causes the largest drop (S1: 95\%$\to$50\%, Overall: 70\%$\to$10\%): without smoothness, the velocity field oscillates between solver evaluations and even short-horizon grasps fail due to imprecise gripper positioning. This confirms that $\mathcal{L}_{\mathrm{vel}}$ is the most critical single component.
Removing $\mathcal{L}_{\mathrm{multistep}}$ has a more targeted effect (Overall: 70\%$\to$20\%): S1 remains moderate (60\%) since the first subtask is shorter, but without trajectory-level feedback the drift becomes fatal for the longer second phase.
Replacing RK4 with Euler reduces overall to 40\%; S1 remains relatively high (80\%), confirming that Euler's $\mathcal{O}(\Delta t)$ error compounds primarily over the longer second phase.
A shared PointNet drops overall to 45\%: a single encoder must simultaneously parse near-field contact geometry and far-field object positions from a merged cloud, degrading both.

\subsection{Discussion}

The performance gap scales with task difficulty: near-saturated in simulation, 15--40 percentage points on short-horizon tasks, and decisive on long-horizon tasks (70\% vs.\ 0\%). This progression follows directly from the train--inference gap analysis in Section~\ref{sec:probdef}: velocity-field drift is manageable over short horizons but compounds to fatal errors for multi-phase tasks where the integration path is much longer. As an indirect consequence, the $S$-step gradient amplification of $\mathcal{L}_{\mathrm{multistep}}$ (analyzed in Section~\ref{sec:multistep}) forces the PointNet encoder to resolve visual details that pointwise losses are insensitive to---an effect validated by Traj-Consistent FM's zero cup-discrimination errors on Cup Stacking. DP3 and Consistency FM represent the diffusion and flow matching paradigms for 3D point cloud policies. We exclude ManiCM \cite{lu2024manicm} and Consistency Policy \cite{prasad2024consistencyfm} because consistency distillation is orthogonal---it compresses inference steps, while we address training and integration of the velocity field. The 0\% overall baseline success on long-horizon tasks directly manifests compounding integration drift, not weak tuning. Both DP3 and Consistency FM are heavily optimized state-of-the-art methods, evidenced by their competitive short-horizon performance (Table~\ref{tab:real_short}) and DP3's 55\% S1 success on Cup Stacking (Table~\ref{tab:real_long}). Because these baselines rely on pointwise supervision, small per-step errors accumulate across long horizons, physically manifesting as phase-transition failures---stalling after a successful grasp or drifting toward incorrect receptacles (Table~\ref{tab:real_long}). Traj-Consistent FM's trajectory-level supervision directly arrests this compounding error.

%%%%%%%%%%%%%%%%%%%%%%%%%%%%%%%%%%%%%%%%%%%%%%%%%%%%%%%%%%%%%%%%%%%%%%%%%%%%%%%%
\section{CONCLUSION}

We presented Trajectory-Consistent Flow Matching, addressing the train--inference gap in flow matching policies through auxiliary rectified flow velocity regression, multi-step trajectory consistency training, velocity smoothness regularization, and RK4 inference. On four real-robot tasks across two platforms, our method achieves 70\% and 60\% overall success on long-horizon multi-phase tasks where both baselines score 0\%, and reaches 100\% on precision tool placement. Our ablation confirms that no single component is sufficient: the interaction between trajectory-level supervision, smooth velocity fields, and higher-order integration is what enables reliable long-horizon execution. The core insight---that trajectory-level supervision is essential when pointwise-trained velocity fields are integrated as ODEs---applies broadly to any iterative generative policy, including diffusion-based formulations. Future work includes learning segment endpoints and step counts adaptively, incorporating optimal transport couplings for improved noise-data pairings, and extension to bimanual and deformable-object manipulation tasks.

%%%%%%%%%%%%%%%%%%%%%%%%%%%%%%%%%%%%%%%%%%%%%%%%%%%%%%%%%%%%%%%%%%%%%%%%%%%%%%%%

\balance
\bibliographystyle{IEEEtran}
\bibliography{references}

@inproceedings{chi2023diffusionpolicy,
  title={Diffusion policy: Visuomotor policy learning via action diffusion},
  author={Chi, Cheng and Feng, Siyuan and Du, Yilun and Xu, Zhenjia and Cousineau, Eric and Burchfiel, Benjamin and Song, Shuran},
  booktitle={Robotics: Science and Systems},
  year={2023}
}

@inproceedings{lipman2023flow,
  title={Flow matching for generative modeling},
  author={Lipman, Yaron and Chen, Ricky T. Q. and Ben-Hamu, Heli and Nickel, Maximilian and Le, Matt},
  booktitle={International Conference on Learning Representations},
  year={2023}
}

@inproceedings{albergo2023cfm,
  title={Building normalizing flows with stochastic interpolants},
  author={Albergo, Michael S and Vanden-Eijnden, Eric},
  booktitle={International Conference on Learning Representations},
  year={2023}
}

@inproceedings{liu2023rectifiedflow,
  title={Flow straight and fast: Learning to generate and transfer data with rectified flow},
  author={Liu, Xingchao and Gong, Chengyue and Liu, Qiang},
  booktitle={International Conference on Learning Representations},
  year={2023}
}

@article{heek2024multistep,
  title={Multistep consistency models},
  author={Heek, Jonathan and Hoogeboom, Emiel and Salimans, Tim},
  journal={arXiv preprint arXiv:2403.06807},
  year={2024}
}

@inproceedings{ze20243ddp,
	title={3D Diffusion Policy: Generalizable Visuomotor Policy Learning via Simple 3D Representations},
	author={Yanjie Ze and Gu Zhang and Kangning Zhang and Chenyuan Hu and Muhan Wang and Huazhe Xu},
	booktitle={Proceedings of Robotics: Science and Systems (RSS)},
	year={2024}
}

@inproceedings{zhao2023act,
  title={Learning fine-grained bimanual manipulation with low-cost hardware},
  author={Zhao, Tony Z and Kumar, Vikash and Levine, Sergey and Finn, Chelsea},
  booktitle={Robotics: Science and Systems},
  year={2023}
}

@inproceedings{florence2021ibc,
  title={Implicit behavioral cloning},
  author={Florence, Pete and Lynch, Corey and Zeng, Andy and Ramirez, Oscar A and Wahid, Ayzaan and Downs, Laura and Wong, Adrian and Lee, Johnny and Mordatch, Igor and Tompson, Jonathan},
  booktitle={Conference on Robot Learning},
  pages={158--168},
  year={2021}
}

@inproceedings{pomerleau1989alvinn,
  title={Alvinn: An autonomous land vehicle in a neural network},
  author={Pomerleau, Dean A},
  booktitle={Advances in Neural Information Processing Systems},
  year={1989}
}

@inproceedings{qi2017pointnet,
  title={Pointnet: Deep learning on point sets for 3d classification and segmentation},
  author={Qi, Charles R and Su, Hao and Mo, Kaichun and Guibas, Leonidas J},
  booktitle={IEEE Conference on Computer Vision and Pattern Recognition},
  pages={652--660},
  year={2017}
}

@article{tong2024cfmot,
  title={Improving and generalizing flow-matching for conditional generation},
  author={Tong, Alexander and Malkin, Nikolay and Fatras, Kilian and Atanackovic, Lazar and Zhang, Yanlei and Huguet, Guillaume and Wolf, Guy and Bengio, Yoshua},
  journal={arXiv preprint arXiv:2302.00482},
  year={2023}
}

@inproceedings{ross2011dagger,
  title={A reduction of imitation learning and structured prediction to no-regret online learning},
  author={Ross, St{\'e}phane and Gordon, Geoffrey and Bagnell, Drew},
  booktitle={Proceedings of the Fourteenth International Conference on Artificial Intelligence and Statistics},
  pages={627--635},
  year={2011}
}

@inproceedings{prasad2024consistencyfm,
  title={Consistency policy: Accelerated visuomotor policies via consistency distillation},
  author={Prasad, Aaditya and Lin, Kevin and Wu, Jimmy and Zhou, Linqi and Bohg, Jeannette},
  booktitle={Robotics: Science and Systems},
  year={2024}
}

@article{black2024pi0,
  title={$\pi_0$: A vision-language-action flow model for general robot control},
  author={Black, Kevin and Brown, Noah and Driess, Danny and Esmail, Adnan and Equi, Michael and Finn, Chelsea and Fusai, Niccolo and Groom, Lachy and Hausman, Karol and Ichter, Brian and others},
  journal={arXiv preprint arXiv:2410.24164},
  year={2024}
}

@inproceedings{butcher2016runge,
  title={Numerical methods for ordinary differential equations},
  author={Butcher, John C},
  publisher={John Wiley \& Sons},
  year={2016}
}

@inproceedings{perez2018film,
  title={{FiLM}: Visual reasoning with a general conditioning layer},
  author={Perez, Ethan and Strub, Florian and De Vries, Harm and Dumoulin, Vincent and Courville, Aaron},
  booktitle={AAAI Conference on Artificial Intelligence},
  year={2018}
}

@inproceedings{yu2020metaworld,
  title={Meta-world: A benchmark and evaluation for multi-task and meta reinforcement learning},
  author={Yu, Tianhe and Quillen, Deirdre and He, Zhanpeng and Julian, Ryan and Hausman, Karol and Finn, Chelsea and Levine, Sergey},
  booktitle={Conference on Robot Learning},
  pages={1094--1100},
  year={2020},
  organization={PMLR}
}

@article{yang2024consistencyflowmatching,
  title={Consistency Flow Matching: Defining Straight Flows with Velocity Consistency},
  author={Yang, Ling and Zhu, Zixiang and Hong, Zhilong and Xu, Minghao and Zhao, Wentao and Li, Hao and Zhang, Wenfei and Zhang, Zhiwei and Cui, Bin and Huang, Gao},
  journal={arXiv preprint arXiv:2407.02398},
  year={2024}
}

@inproceedings{zhang2025flowpolicy,
  title={Flowpolicy: Enabling fast and robust 3d flow-based policy via consistency flow matching for robot manipulation},
  author={Zhang, Qinglun and Liu, Zhen and Fan, Haoqiang and Liu, Guanghui and Zeng, Bing and Liu, Shuaicheng},
  booktitle={Proceedings of the AAAI Conference on Artificial Intelligence},
  volume={39},
  number={14},
  pages={14754--14762},
  year={2025}
}

@article{noh20253d,
  title={3d flow diffusion policy: Visuomotor policy learning via generating flow in 3d space},
  author={Noh, Sangjun and Nam, Dongwoo and Kim, Kangmin and Lee, Geonhyup and Yu, Yeonguk and Kang, Raeyoung and Lee, Kyoobin},
  journal={arXiv preprint arXiv:2509.18676},
  year={2025}
}

@article{chisari2024learning,
  title={Learning robotic manipulation policies from point clouds with conditional flow matching},
  author={Chisari, Eugenio and Heppert, Nick and Argus, Max and Welschehold, Tim and Brox, Thomas and Valada, Abhinav},
  journal={arXiv preprint arXiv:2409.07343},
  year={2024}
}

@article{hu2024adaflow,
  title={Adaflow: Imitation learning with variance-adaptive flow-based policies},
  author={Hu, Xixi and Liu, Qiang and Liu, Xingchao and Liu, Bo},
  journal={Advances in Neural Information Processing Systems},
  volume={37},
  pages={138836--138858},
  year={2024}
}

@inproceedings{esser2024sd3,
  title={Scaling Rectified Flow Transformers for High-Resolution Image Synthesis},
  author={Esser, Patrick and Kulal, Sumith and Blattmann, Andreas and Entezari, Rahim and M{\"u}ller, Jonas and Saini, Harry and Levi, Yam and Lorenz, Dominik and Sauer, Axel and Boesel, Frederic and Podell, Dustin and Dockhorn, Tim and English, Zion and Rombach, Robin},
  booktitle={Proceedings of the 41st International Conference on Machine Learning},
  year={2024}
}

@inproceedings{liu2023instaflow,
  title={{InstaFlow}: One Step is Enough for High-Quality Diffusion-Based Text-to-Image Generation},
  author={Liu, Xingchao and Zhang, Xiwen and Ma, Jianzhu and Peng, Jian and Liu, Qiang},
  booktitle={International Conference on Learning Representations},
  year={2024}
}

@inproceedings{lu2022dpmsolver,
  title={{DPM-Solver}: A Fast {ODE} Solver for Diffusion Probabilistic Model Sampling in Around 10 Steps},
  author={Lu, Cheng and Zhou, Yuhao and Bao, Fan and Chen, Jianfei and Li, Chongxuan and Zhu, Jun},
  booktitle={Advances in Neural Information Processing Systems},
  volume={35},
  year={2022}
}

@article{lu2024manicm,
  title={{ManiCM}: Real-time 3D Diffusion Policy via Consistency Model for Robotic Manipulation},
  author={Lu, Guanxing and Gao, Zifeng and Chen, Tianxing and Ding, Wenbo and Zhang, Jingwei and Wang, Ziwei},
  journal={arXiv preprint arXiv:2406.01586},
  year={2024}
}

@inproceedings{rouxel2024flowil,
  title={Flow Matching Imitation Learning for Multi-Support Manipulation},
  author={Rouxel, Antoine and Rohou, Simon and Kheddar, Abderrahmane},
  booktitle={IEEE/RSJ International Conference on Intelligent Robots and Systems (IROS)},
  year={2024}
}

@article{li2024tpcfm,
  title={Temporal Pair Consistency Guided Rectified Flow},
  author={Li, Guanzhou and Zhang, Pengfei and Liu, Chang-Chieh and Wu, Chun-Guang},
  journal={arXiv preprint arXiv:2501.12540},
  year={2025}
}

@article{kim2024ctm,
  title={Consistency Trajectory Models: Learning Probability Flow ODE Trajectory of Diffusion},
  author={Kim, Dongjun and Lai, Chieh-Hsin and Liao, Wei-Hsiang and Murata, Naoki and Takida, Yuhta and Uesaka, Toshimitsu and He, Yutong and Mitsufuji, Yuki and Ermon, Stefano},
  journal={arXiv preprint arXiv:2310.02279},
  year={2024}
}

@article{zheng2024dpmsolver3,
  title={{DPM-Solver-v3}: Improved Diffusion {ODE} Solver with Empirical Model Statistics},
  author={Zheng, Kaiwen and Lu, Cheng and Chen, Jianfei and Zhu, Jun},
  journal={Advances in Neural Information Processing Systems},
  volume={36},
  year={2024}
}
%%%%%%%%%%%%%%%%%%%%%%%%%%%%%%%%%%%%%%%%%%%%%%%%%%%%%%%%%%%%%%%%%%%%%%%%%%%%%%%%

\end{document}